\documentclass[twocolumn,conference]{IEEEtran}
\usepackage[T1]{fontenc}
\usepackage[latin9]{inputenc}
\usepackage{array}
\usepackage{float}
\usepackage{booktabs}
\usepackage{url}
\usepackage{dsfont}
\usepackage{amsmath}
\usepackage{amssymb}
\usepackage{graphicx}
\PassOptionsToPackage{normalem}{ulem}
\usepackage{ulem}
\usepackage[unicode=true,
 bookmarks=true,bookmarksnumbered=true,bookmarksopen=true,bookmarksopenlevel=1,
 breaklinks=false,pdfborder={0 0 0},pdfborderstyle={},backref=false,colorlinks=false]
 {hyperref}
\hypersetup{pdftitle={Your Title},
 pdfauthor={Your Name},
 pdfpagelayout=OneColumn, pdfnewwindow=true, pdfstartview=XYZ, plainpages=false}

\makeatletter

\providecommand{\tabularnewline}{\\}

\floatstyle{ruled}
\newfloat{algorithm}{tbp}{loa}
\providecommand{\algorithmname}{Algorithm}
\floatname{algorithm}{\protect\algorithmname}

\usepackage{graphicx,epstopdf}
\IEEEoverridecommandlockouts
\usepackage{balance}
\usepackage[caption=false,font=footnotesize]{subfig}
\usepackage{tikz}
\usepackage{amsmath}
\usepackage{amssymb}
\allowdisplaybreaks

\makeatother

\begin{document}
\title{Differentiable Bayesian~Structure~Learning with Acyclicity~Assurance}
\author{\IEEEauthorblockN{Quang-Duy~Tran, Phuoc~Nguyen, Bao~Duong, Thin~Nguyen}\IEEEauthorblockA{\emph{Applied Artificial Intelligence Institute (A}\textsuperscript{\emph{2}}\emph{I}\textsuperscript{\emph{2}}\emph{),
Deakin University, }Geelong, Australia\\
\{\href{mailto:q.tran@deakin.edu.au}{q.tran}, \href{mailto:phuoc.nguyen@deakin.edu.au}{phuoc.nguyen},
\href{mailto:duongng@deakin.edu.au}{duongng}, \href{mailto:thin.nguyen@deakin.edu.au}{thin.nguyen}\}@deakin.edu.au}}
\maketitle
\begin{abstract}
Score-based approaches in the structure learning task are thriving
because of their scalability. Continuous relaxation has been the key
reason for this advancement. Despite achieving promising outcomes,
most of these methods are still struggling to ensure that the graphs
generated from the latent space are acyclic by minimizing a defined
score. There has also been another trend of permutation-based approaches,
which concern the search for the topological ordering of the variables
in the directed acyclic graph in order to limit the search space of
the graph. In this study, we propose an alternative approach for strictly
constraining the acyclicty of the graphs with an integration of the
knowledge from the topological orderings. Our approach can reduce
inference complexity while ensuring the structures of the generated
graphs to be acyclic. Our empirical experiments with simulated and
real-world data show that our approach can outperform related Bayesian
score-based approaches.

\end{abstract}


\section{Introduction\label{sec:Introduction}}

Structure learning aims to uncover the underlying directed acyclic
graphs~(DAGs) from observational data that can represent statistical
or causal relationships between variables. The structure learning
task has many applications in biology~\cite{Sachs_etal_05Causal},
economics~\cite{Koller_Friedman_09Probabilistic}, and interpretable
machine learning~\cite{Molnar_etal_20Interpretable}. Correspondingly,
it is gaining scientific interest in various domains such as computer
science, statistics, and bioinformatics~\cite{Squires_Uhler_22Causal}.
One challenge of traditional structure learning methods such as GES~\cite{Chickering_02Optimal}
is the combinatorial search space of possible DAGs~\cite{chickering2004large}.
NO-TEARS~\cite{Zheng_etal_18DAGs} proposes a solution for this challenge
by relaxing the formulation of the learning task in a continuous space
and employs continuous optimization techniques. However, with the
continuous representations, another challenge also arises which is
the acyclicity constraint of the graphs. 

In most continuous score-based methods~\cite{Zheng_etal_18DAGs,zheng2020learning,Lorch_etal_21DiBS,Lorch_etal_22Amortized},
the constraints of graph acyclicity are defined in a form of a penalizing
score and minimizing the score will also minimize the cyclicity of
the graphs. This type of approach requires a large number of running
steps with complex penalization weight scheduling to ensure the correctness
of the constraint, which varies greatly depending on settings. This
lack of certainty will affect the quality and restrict the applicability
of the learned structures. Another approach is to embed the constraint
acyclicity in the generative model of the graphs such as in \cite{Cundy_etal_21BCD}
by utilizing weighted adjacency matrices that can be decomposed into
the combinations of a permutation matrix and a strictly lower triangular
matrix. Our study is inspired by this approach by using a direct constraint
in the generation process instead of a post-hoc penalizing score.

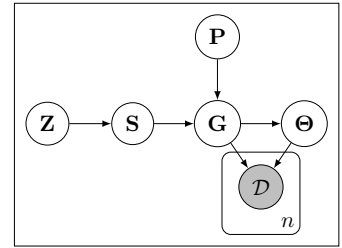
\begin{figure}[t]
\noindent \begin{centering}
\begin{centering}
\resizebox{0.5\linewidth}{!}{\begin{tikzpicture}  
	\usetikzlibrary {shapes.geometric,calc}
	\node [circle, draw] (Z) at (0, 0) {$\mathbf{Z}$};   
	\node [circle, draw] (S) at ($(Z.east)+(1, 0)$) {$\mathbf{S}$};
	\node [circle, draw] (G) at ($(S.east)+(1, 0)$) {$\mathbf{G}$}; 
	\node [circle, draw] (P) at ($(G.north)+(0, 1)$) {$\mathbf{P}$}; 
	\node [circle, draw] (T) at ($(G.east)+(1, 0)$) {$\mathbf{\Theta}$};
	\node [circle, draw, fill=gray!50!white] (D) at ($0.5*(G.east)+0.5*(T.west)+(0,-1)$) {$\mathcal{D}$};
	\node [rectangle, draw, minimum width=3.5em, minimum height=3.7em, rounded corners] (box) at ($(D)+(0,-0.1)$) {};
	\node [] (label) at ($(box.south east)+(-0.2, 0.2)$) {$n$};
	\draw [-latex] (Z) -- (S);
	\draw [-latex] (S) -- (G);
	\draw [-latex] (P) -- (G);
	\draw [-latex] (G) -- (T);
	\draw [-latex] (G) -- (D);
	\draw [-latex] (T) -- (D);
	\path [use as bounding box, draw=black] ([shift={(-0.5em,-0.5em)}]current bounding box.south west) rectangle ([shift={(0.5em,0.5em)}]current bounding box.north east);
\end{tikzpicture}}
\end{centering}
\par\end{centering}
\caption{\label{fig:gen_model}The proposed generative model of the Bayesian
networks in Topological Ordering in Differentiable Bayesian Structure
Learning with Acyclicity Assurance~(TOBAC). $\mathbf{S}$, which
has a strictly upper triangular form, is the adjacency matrix of the
DAG when the topological ordering of the variables is correct. This
matrix is generated from a latent variable $\mathbf{Z}$. For each
instance of the permutation matrix $\mathbf{P}$, the rows and columns
in $\mathbf{S}$ can be permuted to generate an isomorphic graph $\mathbf{G}$.
The variable $\mathbf{\Theta}$ defines the parameters of the local
conditional distributions of the nodes given their parents in $\mathbf{G}$.
The observational data $\mathcal{D}$ consisting of $n$ observations
is assumed to be generated from this generative model.}
\end{figure}

There is a parallel branch of permutation-based causal discovery approaches
whose methods allow us to find the topological ordering in polynomial
time \cite{Chen_etal_19Causal,Gao_etal_20Polynomial,Rolland_etal_22Score,Sanchez_etal_22Diffusion},
which can provide beneficial information. Inspired by these approaches,
we propose a framework, \uline{T}opological \uline{O}rdering
in Differentiable \uline{B}ayesian Structure Learning with \uline{AC}yclicity
Assurance~(TOBAC), to greatly reduce the difficulty of the acyclicity-constraining
task. Conditional inference is performed in this framework with the
condition being the prior knowledge provided from the topological
orderings. Our work is based on the independent factorization property
of a DAG's adjacency matrix into a permutation matrix $\mathbf{P}$
and a strictly upper triangular matrix $\mathbf{S}$ which represents
the adjacency matrix when the ordering is correct. The factorization
$p\left(\mathbf{G},\mathbf{S},\mathbf{P}\right)=p\left(\mathbf{S}\right)p\left(\mathbf{P}\right)p\left(\mathbf{G}\mid\mathbf{S},\mathbf{P}\right)$
allows us to infer $\mathbf{P}$ and $\mathbf{S}$ independently.
Especially, decoupling these enables us to apply recent advances in
learning of topological ordering and probabilistic model inference
techniques. For each case of the permutation matrix $\mathbf{P}$,
we can infer the DAG's strictly upper triangular matrix $\mathbf{S}$
and compute the adjacency of an isomorphic DAG with $\mathbf{G}=\mathbf{P}\mathbf{S}\mathbf{P}^{\top}$.
In order to infer this DAG $\mathbf{G}$, we choose the recent graph
inference approach in this field, DiBS~\cite{Lorch_etal_21DiBS},
as ours inference engine for $\mathbf{S}$. We run experiments on
synthetic data and a real flow cytometry dataset~\cite{Sachs_etal_05Causal}
in linear and nonlinear Gaussian settings. The proposed constraint
in the structure approach shows better DAG predictions and achieves
better performance compared to other approaches.

\noindent \textbf{Contributions. }The main contributions of this study
are summarized as follows
\begin{enumerate}
\item We address the limitations of post-hoc acyclicity constraint scores
by strictly constraining the generative structure of the graphs. By
utilizing the permutation-based decomposition of the adjacency matrix,
we can strictly guarantee the acyclicity constraint in Bayesian network.
\item We introduce TOBAC, a framework for independently inferring and conditioning
on the topological ordering. Our inference process guarantees the
acyclicity of inferred graphs as well as reduces the inference complexity
of the adjacency matrices.
\item We demonstrate the effectiveness of TOBAC in comparison with related
state-of-the-art Bayesian score-based methods on both synthetic and
real-world. Our approach obtains better performance on synthetic linear
and nonlinear Gaussian data and on the real flow cytometry dataset.
\end{enumerate}

\section{Related Works\label{sec:RelatedWorks}}

\paragraph{Bayesian structure learning methods}

Most structure learning approaches can be categorized by their learning
objectives into two main categories. The majority of methods~\cite{Chen_etal_19Causal,Chickering_02Optimal,lachapelle2019gradient,montagna2022scalable,Rolland_etal_22Score,Sanchez_etal_22Diffusion,Zheng_etal_18DAGs,zheng2020learning}
fall into the first category where each method aims to recover a point
estimate or its Markov equivalence class. In the other category, which
is called Bayesian structure learning, the methods~\cite{Cundy_etal_21BCD,Deleu_etal_22Bayesian,Friedman_etal_99Data,Koivisto_06Advances,Lorch_etal_21DiBS,Lorch_etal_22Amortized,Madigan_etal_95Bayesian}
aim to learn the posterior distribution over the DAGs given the observational
data, i.e., $p\left(G\mid\mathcal{D}\right)$. These methods can quantify
the epistemic uncertainty in cases such as limited number of sample
size or non-identifiable models in causal discovery. Previous approaches
to learning the posterior distribution are diverse for example, using
Markov chain Monte Carlo (MCMC)~\cite{Madigan_etal_95Bayesian},
bootstrapping~\cite{Friedman_etal_99Data} with PC~\cite{Spirtes_etal_00Causation}
and GES~\cite{Chickering_02Optimal}, and exact methods with dynamic
programming~\cite{Koivisto_06Advances}. Recent Bayesian structure
learning approaches use more advanced methods such as variational
inference~\cite{Cundy_etal_21BCD,Lorch_etal_21DiBS,Lorch_etal_22Amortized}
or Generative Flow Networks (GFlowNets)~\cite{Bengio_etal_22GFlowNet,Deleu_etal_22Bayesian,Nishikawa_etal_22Bayesian,Deleu_etal_23Joint}.

\paragraph{Permutation-based methods}

Searching over the space of the permutations of the variables is significantly
faster than searching over the space of possible DAGs. When the correct
topological ordering of a DAG is found, a skeleton containing possible
relations can be constructed and the DAG can be easily retrieved from
this skeleton using the available conditional independence tests~\cite{buhlmann2014cam,raskutti2018learning,solus2021consistency,wang2017permutation,squires2020permutation}.
There has been many approaches for both linear~\cite{Chen_etal_19Causal,zhao2022learning}
and nonlinear additive noise models~\cite{Gao_etal_20Polynomial,Rolland_etal_22Score,Sanchez_etal_22Diffusion,montagna2022scalable}.
EqVar~\cite{Chen_etal_19Causal} learns the topological orderings
with the assumption of equal variances. NPVar~\cite{Gao_etal_20Polynomial}
extends the work of EqVar by replacing the error variances with the
corresponding residual variances and modeling using the nonparametric
setting. Both of these methods iteratively find the root nodes to
the leaf nodes of the causal DAG. Alternatively, SCORE~\cite{Rolland_etal_22Score}
finds the leaves from a score estimated to match the gradient of the
log probability distribution of the variables.

\paragraph{Score-based methods}

For fully observed data, early method such as GES~\cite{Chickering_02Optimal}
uses greedy search to efficiently search for the causal order in the
permutation space by maximizing a score function. Recent developments
efficiently search for the sparsest permutation, as in the Sparsest
Permutation algorithm~\cite{raskutti2018learning} and Greedy Sparsest
Permutation algorithm~\cite{solus2021consistency}, over the vertices
of a permutohedron representing the space of permutations of DAGs,
and reduce the search space by contracting the vertices corresponding
to the same DAG~\cite{solus2021consistency}. These methods are also
adapted for interventional data in~\cite{wang2017permutation,squires2020permutation}.
Another recent approach in this category is the group of DAG-GFlowNet~\cite{Deleu_etal_22Bayesian,Nishikawa_etal_22Bayesian,Deleu_etal_23Joint}
methods, which utilizes GFlowNet to search over the states of DAGs
and can approximate the posterior distribution using both observational
and interventional data.

\paragraph{Continuous score-based methods}

Structure learning approaches with continuous relaxation~\cite{zheng2020learning,lachapelle2019gradient,Yu_etal_19DAG,Cundy_etal_21BCD,Lorch_etal_21DiBS,Lorch_etal_22Amortized}
is developing rapidly since NO-TEARS~\cite{Zheng_etal_18DAGs} was
introduced. A continuous search space allows us to optimize or infer
using gradient-based approaches to avoid search over the large space
of discrete DAGs. Bayesian inference with variational inference is
one category of the gradient-based approaches used in the latest frameworks~\cite{Cundy_etal_21BCD,Lorch_etal_21DiBS,Lorch_etal_22Amortized}.
BCD~Nets~\cite{Cundy_etal_21BCD} decomposes the weighted adjacency
matrix to a permutation matrix and a strictly lower triangular matrix,
and infers the probabilities of these matrices using the evidence
lower bound (ELBO) of the variational inference problem. DiBS~\cite{Lorch_etal_21DiBS}
models the probabilities of the edges using a bilinear generative
model from the latent space and infers the posterior using Stein variational
gradient descent. 

\paragraph{Topological ordering in continuous score-based methods}

Topological orderings appear variously in these continuous approaches.
These recent advances use probabilistic models based on neural networks
for (1) estimating the topological order directly as in BCD~Nets~\cite{Cundy_etal_21BCD}
in the form of a permutation matrix, or (2) sidestepping the topological
order estimation by including an auxiliary DAG constraint~\cite{Zheng_etal_18DAGs,zheng2020learning,Lee_etal_19Scaling}
or regularization~\cite{Yu_etal_19DAG,lachapelle2019gradient,Lorch_etal_21DiBS,Lorch_etal_22Amortized}
instead. Our framework belongs to the former category, but we avoid
the complexity in the joint inference of the permutation matrix and
the graph. In this study, we design an acyclicity-ensuring conditional
inference process with the topological ordering as a condition. With
this process setting, we can effectively integrate the knowledge from
the topological ordering into the inference to achieve higher inference
efficiency.

\section{TOBAC: Topological~Ordering in Differentiable Bayesian~Structure~Learning
with Acyclicity~Assurance\label{sec:Methodology}}

\subsection{Acyclicity Assurance via Decomposition of Adjacency Matrix\label{subsec:Intuition}}

To analyze the decomposition of the adjacency matrix of a directed
acyclic graph (DAG), we need to start from the topological orderings.
A topological ordering (or, in short, an ordering) is a topological
sort of the variables in a DAG. Let $\mathbf{\pi}=\left[\pi_{1},\pi_{2},\dots,\pi_{d}\right]$
be the corresponding ordering of the variables $\mathbf{X}=\left[X_{1},X_{2},\dots,X_{d}\right]$,
and $\pi_{i}<\pi_{j}$ means that $X_{i}\in nd\left(X_{j}\right)$
where $nd\left(X_{j}\right)$ is the set containing the non-descendants
of $X_{j}$. We consider the canonical case where the ordering of
the variables is already correct, i.e., $\mathbf{\pi}^{*}=\left[\pi_{i}^{*}=i,\forall i\right]$.
In this case, because every variable $X_{i}$ is a non-descendant
of $X_{j}$ if $i<j$, the adjacency matrix $\mathbf{G}$ of the DAG
will become a strictly upper triangular matrix $\mathbf{S}\in\left\{ 0,1\right\} ^{d\times d}$. 

In order to generalize to any ordering, we use a permutation matrix
$\mathbf{P}$ that transforms the ordering $\pi$ to $\pi^{*}$ as
\begin{equation}
P_{i,j}=\begin{cases}
1 & \textrm{if \ensuremath{i=\pi_{j}},}\\
0 & \textrm{otherwise.}
\end{cases}\label{eq:perm_matrix}
\end{equation}
With this permutation matrix, we can always generate the adjacency
matrix of an isomorphic DAG by
\begin{equation}
\mathbf{G}=\mathbf{P}\mathbf{S}\mathbf{P}^{\top}.\label{eq:adj_decompose}
\end{equation}
This formulation shifts the corresponding rows and columns in $\mathbf{G}$
from the canonical ordering \textbf{$\mathbf{\pi}^{*}$} to the ordering
$\mathbf{\pi}$. As the canonical adjacency matrix is acyclic, the
derived graphs will always satisfy acyclicity. By employing this decomposition
in the generative process, the acyclicity of the inferred graphs will
always be satisfied.

\subsection{Representing the Canonical Adjacency Matrix in a Latent~Space\label{subsec:LatentRepresentation}}

With every permutation matrix $\mathbf{P}$, we only need to find
the equivalent canonical adjacency matrix $\mathbf{S}$. Following
the DiBS approach in~\cite{Lorch_etal_21DiBS}, this matrix is sampled
from a latent variable \textbf{$\mathbf{Z}$} consisting of two embedding
matrices $\mathbf{U}=\left[\mathbf{u}_{1},\mathbf{u}_{2},\ldots,\mathbf{u}_{d-1}\right],\mathbf{u}_{i}\in\mathbb{R}^{k}$
and $\mathbf{V}=\left[\mathbf{v}_{1},\mathbf{v}_{2},\ldots,\mathbf{v}_{d-1}\right],\mathbf{v}_{j}\in\mathbb{R}^{k}$.
Due to the nature of strictly upper triangular matrices, only $d-1$
vectors in each embedding are needed to construct $\mathbf{S}$ instead
of $d$ vectors as in DiBS. Following this configuration, the dimension
$k$ of the latent vectors is chosen to be greater or equal to $d-1$
to ensure that the generated graphs are not constrained in rank. We
represent the probabilities of values in $\mathbf{S}$ as follows
\begin{align}
\mathbf{S_{\alpha}}\left(\mathbf{Z}\right)_{i,j} & :=p_{\alpha}\left(S_{i,j}=1\mid\mathbf{u}_{i},\mathbf{v}_{j-1}\right)\label{eq:bilinear_prob}\\
 & =\begin{cases}
\sigma_{\alpha}\left(\mathbf{u}_{i}^{\top}\mathbf{v}_{j-1}\right) & \textrm{if \ensuremath{j>i},}\\
0 & \textrm{otherwise;}
\end{cases}\label{eq:bilinear_alpha}
\end{align}
where $\sigma_{\alpha}\left(x\right)=1/\left(1+\exp\left(-\alpha x\right)\right)$
and the term $\alpha$ will be increased each step to make the sigmoid
function $\sigma_{\alpha}\left(x\right)$ converge to the Heaviside
step function $\mathds{1}\left[x>0\right]$. As $\alpha\rightarrow\infty$,
the converged generated $\mathbf{S}_{\infty}$ will become
\begin{equation}
\mathbf{S}_{\infty}\left(\mathbf{Z}\right)_{i,j}:=\begin{cases}
\mathcal{\mathds{1}}\left[\mathbf{u}_{i}^{\top}\mathbf{v}_{j-1}>0\right] & \textrm{if \ensuremath{j>i},}\\
0 & \textrm{otherwise.}
\end{cases}\label{eq:bilinear_infty}
\end{equation}
The probability of the edges in the adjacency matrix $\mathbf{G}$
given the latent variable $\mathbf{Z}$ and the permutation matrix
$\mathbf{P}$ can be computed by
\begin{equation}
\begin{aligned} & p_{\alpha}\left(G_{i,j}=1\mid\mathbf{Z},\mathbf{P}\right)\\
 & :=\sum_{a=1}^{d}\sum_{b=1}^{d}P_{i,b}p_{\alpha}\left(S_{b,a}=1\mid\mathbf{Z}\right)P_{a,j}^{\top}
\end{aligned}
\label{eq:prob_g_z_p}
\end{equation}

\subsection{Estimating the Latent Variable using Bayesian~Inference\label{subsec:LatentEstimation}}

In order to estimate the latent variable $\mathbf{Z}$, extending
the approach from~\cite{Lorch_etal_21DiBS}, we first consider the
generative model given in Figure~\ref{fig:gen_model}. In this figure,
a Bayesian network consists of a pair of variables $\left(\mathbf{G},\mathbf{\Theta}\right)$
where $\mathbf{\Theta}$ defines the parameters of the local conditional
distributions at each variable given its parents in the DAG. This
generative model are assumed to generate the observational data $\mathcal{D}$
containing $n$ observations. 

Given a permutation matrix $\mathbf{P}$, the generative model conditioned
on $\mathbf{P}$ can be factorized as
\begin{equation}
\begin{aligned} & p\left(\mathbf{Z},\mathbf{S},\mathbf{G},\mathbf{\Theta},\mathcal{D}\mid\mathbf{P}\right)\\
 & =p\left(\mathbf{Z}\right)p\left(\mathbf{S}\mid\mathbf{Z}\right)p\left(\mathbf{G}\mid\mathbf{S},\mathbf{P}\right)p\left(\mathbf{\Theta}\mid\mathbf{G}\right)p\left(\mathcal{D}\mid\mathbf{G},\mathbf{\Theta}\right).
\end{aligned}
\label{eq:gen_factorize}
\end{equation}
For any function $f\left(\mathbf{G},\mathbf{\Theta}\right)$ of interest,
we can compute its expectation from the distribution $p\left(\mathbf{G},\mathbf{\Theta}\mid\mathcal{D},\mathbf{P}\right)$
by inferring $p\left(\mathbf{Z},\mathbf{\Theta}\mid\mathcal{D},\mathbf{P}\right)$
with the following formula
\begin{equation}
\begin{aligned} & \mathbb{E}_{p\left(\mathbf{G},\mathbf{\Theta}\mid\mathcal{D},\mathbf{P}\right)}\left[f\left(\mathbf{G},\mathbf{\Theta}\right)\right]\\
 & =\mathbb{E}_{p\left(\mathbf{Z},\mathbf{\Theta}\mid\mathcal{D},\mathbf{P}\right)}\left[\frac{\mathbb{E}_{p\left(\mathbf{G}\mid\mathbf{Z},\mathbf{P}\right)}\left[f\left(\mathbf{G},\mathbf{\Theta}\right)p\left(\mathbf{\Theta},\mathcal{D}\mid\mathbf{G}\right)\right]}{\mathbb{E}_{p\left(\mathbf{G}\mid\mathbf{Z},\mathbf{P}\right)}\left[p\left(\mathbf{\Theta},\mathcal{D}\mid\mathbf{G}\right)\right]}\right],
\end{aligned}
\label{eq:expectation}
\end{equation}
where $p\left(\mathbf{\Theta},\mathcal{D}\mid\mathbf{G}\right)=p\left(\mathbf{\Theta}\mid\mathbf{G}\right)p\left(\mathcal{D}\mid\mathbf{G},\mathbf{\Theta}\right)$
and $p\left(\mathbf{G}\mid\mathbf{Z},\mathbf{P}\right)$ is computed
using a graph prior (e.g., Erd\H{o}s\textendash R\'{e}nyi~\cite{Erdos_Renyi_60Evolution}
or scale-free~\cite{Barabasi_Albert_99Emergence}) with the soft
graph in Equation (\ref{eq:prob_g_z_p}). The function $f\left(\mathbf{G},\mathbf{\Theta}\right)$
in Equation~(\ref{eq:expectation}) acts as a placeholder for $p\left(\mathcal{D}\mid\mathbf{G},\mathbf{\Theta}\right)$
in this study or any other functions depending on the setting of each
structure learning task. The distributions of the parameters $p\left(\mathbf{\Theta}\mid\mathbf{G}\right)$
and the data $p\left(\mathcal{D}\mid\mathbf{G},\mathbf{\Theta}\right)$
are chosen differently for the linear and nonlinear Gaussian models.
In the linear Gaussian model, the log probability of the parameters
given the graph is
\begin{equation}
\log p\left(\mathbf{\Theta}\mid\mathbf{G}\right)=\sum_{i,j}G_{i,j}\log\mathcal{N}\left(\theta_{i,j};\mu_{e},\sigma_{e}^{2}\right),
\end{equation}
where $\mu_{e}$ and $\sigma_{e}$ are the mean and standard deviation
of the Gaussian edge weights, and the log likelihood is as follows
\begin{equation}
\log p\left(\mathcal{D}\mid\mathbf{G},\mathbf{\Theta}\right)=\sum_{i=1}^{d}\log\mathcal{N}\left(X_{i};\mathbf{\theta}_{i}^{\top}\mathbf{X}_{\mathrm{pa}\left(i\right)},\sigma_{obs}^{2}\right),
\end{equation}
where $\sigma_{obs}$ is the standard deviation of the additive observation
noise at each node. In the nonlinear Gaussian model, we follow \cite{zheng2020learning,Lorch_etal_21DiBS}
by using the feed-forward neural networks (FFNs) denoted by $\mathrm{FFN}\left(\cdot;\mathbf{\Theta}\right):\mathbb{R}^{d}\rightarrow\mathbb{R}$
to represent the relation of the variables with their parents (i.e.,
$f_{i}\left(\mathbf{X}_{pa\left(i\right)}\right)$ for each $X_{i}$).
For each variable $X_{i}$, the output is computed as
\begin{equation}
\begin{aligned} & \mathrm{FFN}\left(\mathbf{x};\mathbf{\Theta}^{\left(i\right)}\right)\\
 & :=\mathbf{\Theta}^{\left(i,L\right)}f_{\mathrm{a}}\left(\dots\mathbf{\Theta}^{\left(i,2\right)}f_{\mathrm{a}}\left(\mathbf{\Theta}^{\left(i,1\right)}\mathbf{x}+\mathbf{\theta}_{b}^{\left(i,1\right)}\right)+\mathbf{\theta}_{b}^{\left(i,2\right)}\dots\right)\\
 & \quad+\mathbf{\theta}_{b}^{\left(i,L\right)},
\end{aligned}
\end{equation}
where $\mathbf{\Theta}^{(i,l)}\in\mathbb{R}^{d_{l}\times d_{l-1}}$
is the weight matrix, $\mathbf{\theta}_{b}^{\left(i,l\right)}\in\mathbb{R}^{d_{l}}$
is the bias vector, and $f_{\mathrm{a}}$ is the activation function.
From the model in this setting, the log probability of the parameters
is 
\begin{equation}
\begin{aligned}\log p\left(\mathbf{\Theta}\mid\mathbf{G}\right) & =\sum_{i=1}^{d}\Bigg(\sum_{a=1}^{d_{1}}\bigg(\log\mathcal{N}\left(\left(\mathbf{\theta}_{b}^{\left(i,1\right)}\right)_{a};0,\sigma_{p}^{2}\right)\\
 & \quad+\sum_{b=1}^{d}G_{i,b}^{\top}\log\mathcal{N}\left(\mathbf{\Theta}_{a,b}^{\left(i,1\right)};0,\sigma_{p}^{2}\right)\bigg)\\
 & \quad+\sum_{l=2}^{L}\sum_{a=1}^{d_{l}}\bigg(\log\mathcal{N}\left(\left(\mathbf{\theta}_{b}^{(i,l)}\right)_{a};0,\sigma_{p}^{2}\right)\\
 & \quad+\sum_{b=1}^{d_{l-1}}\log\mathcal{N}\left(\mathbf{\Theta}_{a,b}^{\left(i,l\right)};0,\sigma_{p}^{2}\right)\bigg)\Bigg),
\end{aligned}
\end{equation}
where $\sigma_{p}$ is the standard deviation of the Gaussian parameters.
In the nonlinear Gaussian setting, the value of $f_{i}\left(\mathbf{X}_{pa\left(i\right)}\right)$
is assumed to be the mean of the distribution of each variable $X_{i}$.
As a result, the log likelihood of this model is as follows
\begin{equation}
\begin{aligned} & \log p\left(\mathcal{D}\mid\mathbf{G},\mathbf{\Theta}\right)\\
 & =\sum_{i=1}^{d}\log\mathcal{N}\left(X_{i};\mathrm{FFN}\left(\mathbf{G}_{i}^{\top}\circ\mathbf{X};\mathbf{\Theta}^{(i)}\right),\sigma_{obs}^{2}\right),
\end{aligned}
\end{equation}
 where ``$\circ$'' denotes the element-wise multiplication.

\subsection{Particle Variational Inference of Intractable~Posterior\label{subsec:SVGD}}

\begin{algorithm*}
\caption{\label{alg:SVGD}SVGD algorithm for inference of $p\left(\mathbf{G},\mathbf{\Theta}\mid\mathcal{D},\mathbf{P}\right)$}

\textbf{Input:} Initial set of latent and parameter particles $\left\{ \left(\mathbf{Z}_{0}^{\left(m\right)},\mathbf{\Theta}_{0}^{\left(m\right)}\right)\right\} _{m=1}^{M}$,
kernel $k$, schedules for $\eta_{t}$, $\alpha_{t}$, observational
data $\mathcal{D}$, and permutation matrix $\mathbf{P}$

\textbf{Output:} Set of adjacency matrices and parameter particles
$\left\{ \left(\mathbf{G}^{\left(m\right)},\mathbf{\Theta}^{\left(m\right)}\right)\right\} _{m=1}^{M}$
\begin{enumerate}
\item Incorporate prior belief of $p\left(\mathbf{G}\mid\mathbf{Z},\mathbf{P}\right)$
into $p\left(\mathbf{Z}\right)$
\item \textbf{for} iteration $t=0$ to $T-1$ \textbf{do}
\item \quad{}Estimate score $\nabla_{\mathbf{Z}}\log p\left(\mathbf{Z},\mathbf{\Theta}\mid\mathcal{D},\mathbf{P}\right)$
given in Equation~(\ref{eq:grad_z}) for every $\mathbf{Z}_{t}^{\left(m\right)}$
\item \quad{}Estimate score $\nabla_{\mathbf{\Theta}}\log p\left(\mathbf{Z},\mathbf{\Theta}\mid\mathcal{D},\mathbf{P}\right)$
given in Equation~(\ref{eq:grad_theta}) for every $\mathbf{\Theta}_{t}^{\left(m\right)}$
\item \quad{}\textbf{for} particle $m=1$ to $M$ \textbf{do}
\item \qquad{}$\mathbf{Z}_{t+1}^{\left(m\right)}\leftarrow\mathbf{Z}_{t}^{\left(m\right)}+\eta_{t}\phi_{t}^{\mathbf{Z}}\left(\mathbf{Z}_{t}^{\left(m\right)},\mathbf{\Theta}_{t}^{\left(m\right)}\right)$,
($\phi_{t}^{\mathbf{Z}}\left(\mathbf{Z}_{t}^{\left(m\right)},\mathbf{\Theta}_{t}^{\left(m\right)}\right)$
from Equation~(\ref{eq:update_z}))
\item \qquad{}$\mathbf{\Theta}_{t+1}^{\left(m\right)}\leftarrow\mathbf{\Theta}_{t}^{\left(m\right)}+\eta_{t}\phi_{t}^{\mathbf{\Theta}}\left(\mathbf{Z}_{t}^{\left(m\right)},\mathbf{\Theta}_{t}^{\left(m\right)}\right)$,
($\phi_{t}^{\mathbf{\Theta}}\left(\mathbf{Z}_{t}^{\left(m\right)},\mathbf{\Theta}_{t}^{\left(m\right)}\right)$
from Equation~(\ref{eq:update_theta}))
\item \textbf{return} $\left\{ \left(\mathbf{G}_{\infty}^{\left(m\right)}=\mathbf{P}\mathbf{S}_{\infty}\left(\mathbf{Z}_{T}^{\left(m\right)}\right)\mathbf{P}^{\top},\mathbf{\Theta}_{T}^{\left(m\right)}\right)\right\} _{m=1}^{M}$
\end{enumerate}
\end{algorithm*}

The joint posterior distribution $p\left(\mathbf{Z},\mathbf{\Theta}\mid\mathcal{D},\mathbf{P}\right)$
is intractable. Stein variational gradient descent~(SVGD)~\cite{Liu_Wang_16Stein,Lorch_etal_21DiBS}
is a suitable method to approximate this joint posterior density due
to its gradient-based approach. The SVGD algorithm iteratively transports
a set of particles to match the target distribution similar to the
gradient descent algorithm in optimization. 

From the proposed generative model, we need to infer the log joint
posterior density of $\mathbf{Z}$ and $\mathbf{\Theta}$ using the
corresponding gradient to variable $\mathbf{Z}$ given by
\begin{equation}
\begin{aligned} & \nabla_{\mathbf{Z}}\log p\left(\mathbf{Z},\mathbf{\Theta}\mid\mathcal{D},\mathbf{P}\right)\\
 & =\nabla_{\mathbf{Z}}\log p\left(\mathbf{Z}\right)+\frac{\nabla_{\mathbf{Z}}\mathbb{E}_{p\left(\mathbf{G}\mid\mathbf{Z},\mathbf{P}\right)}\left[p\left(\mathbf{\Theta},\mathcal{D}\mid\mathbf{G}\right)\right]}{\mathbb{E}_{p\left(\mathbf{G}\mid\mathbf{Z},\mathbf{P}\right)}\left[p\left(\mathbf{\Theta},\mathcal{D}\mid\mathbf{G}\right)\right]}.
\end{aligned}
\label{eq:grad_z}
\end{equation}
The log latent prior distribution $\log p\left(\mathbf{Z}\right)$
is chosen as
\begin{equation}
\begin{aligned}\log p\left(\mathbf{Z}\right) & :=\sum_{i,j}\mathcal{\log N}\left(U_{ij};0,\sigma_{z}^{2}\right)+\sum_{i,j}\mathcal{\log N}\left(V_{ij};0,\sigma_{z}^{2}\right)\\
 & \quad+\mathbb{E}_{p\left(\mathbf{G}\mid\mathbf{Z},\mathbf{P}\right)}\left[\log p\left(\mathbf{G}\mid\mathbf{Z},\mathbf{P}\right)\right]-C,
\end{aligned}
\label{eq:log_latent_prior}
\end{equation}
where $C$ is the log partitioning constant. Similarly, the gradient
corresponding to variable $\mathbf{\Theta}$ is given by

\begin{equation}
\nabla_{\mathbf{\Theta}}\log p\left(\mathbf{Z},\mathbf{\Theta}\mid\mathcal{D},\mathbf{P}\right)=\frac{\nabla_{\mathbf{\Theta}}\mathbb{E}_{p\left(\mathbf{G}\mid\mathbf{Z},\mathbf{P}\right)}\left[p\left(\mathbf{\Theta},\mathcal{D}\mid\mathbf{G}\right)\right]}{\mathbb{E}_{p\left(\mathbf{G}\mid\mathbf{Z},\mathbf{P}\right)}\left[p\left(\mathbf{\Theta},\mathcal{D}\mid\mathbf{G}\right)\right]}.\label{eq:grad_theta}
\end{equation}
The numerator of the second term in Equation~(\ref{eq:grad_z}) can
be approximated using the Gumbel-softmax trick~\cite{Jang_etal_17Categorical,Maddison_etal_17Concrete}
as follows
\begin{equation}
\begin{aligned} & \nabla_{\mathbf{\Theta}}\mathbb{E}_{p\left(\mathbf{G}\mid\mathbf{Z},\mathbf{P}\right)}\left[p\left(\mathbf{\Theta},\mathcal{D}\mid\mathbf{G}\right)\right]\\
 & \approx\mathbb{E}_{p\left(\mathbf{L}\right)}\left[\nabla_{\mathbf{G}}p\left(\mathbf{\Theta},\mathcal{D}\mid\mathbf{G}\right)\Big\vert_{\mathbf{G}=\tilde{\mathbf{G}}_{\tau}\left(\mathbf{L},\mathbf{Z}\right)}\cdot\nabla_{\mathbf{Z}}\tilde{\mathbf{G}}_{\tau}\left(\mathbf{L},\mathbf{Z}\right)\right],
\end{aligned}
\label{eq:gumbel_approx}
\end{equation}
where $\mathbf{L}\sim\textrm{Logistic}\left(0,1\right)^{\left(d-1\right)\times\left(d-1\right)}$
and the matrix $\tilde{\mathbf{G}}_{\tau}\left(\mathbf{L},\mathbf{Z}\right)=\mathbf{P}\tilde{\mathbf{S}}_{\tau}\left(\mathbf{L},\mathbf{Z}\right)\mathbf{P}^{\top}$.
The element-wise definition of $\tilde{\mathbf{S}}_{\tau}$ is
\begin{equation}
\tilde{\mathbf{S}}_{\tau}\left(\mathbf{L},\mathbf{Z}\right)_{i,j}:=\begin{cases}
\sigma_{\tau}\left(L_{i,j-1}+\alpha\mathbf{u}_{i}^{\top}\mathbf{v}_{j-1}\right) & \textrm{if \ensuremath{j>i},}\\
0 & \textrm{otherwise.}
\end{cases}\label{eq:bilinear_gumbel}
\end{equation}
In our experiments, we choose $\tau=1$ in accordance with~\cite{Lorch_etal_21DiBS}.

The kernel for SVGD proposed in~\cite{Lorch_etal_21DiBS} is 
\begin{equation}
\begin{aligned} & k\left(\left(\mathbf{Z},\mathbf{\Theta}\right),\left(\mathbf{Z}^{\prime},\mathbf{\Theta}^{\prime}\right)\right)\\
 & :=\exp\left(-\frac{1}{\gamma_{z}}\left\Vert \mathbf{Z}-\mathbf{Z}^{\prime}\right\Vert _{2}^{2}\right)+\exp\left(-\frac{1}{\gamma_{\theta}}\left\Vert \mathbf{\Theta}-\mathbf{\Theta}^{\prime}\right\Vert _{2}^{2}\right).
\end{aligned}
\label{eq:svgd_kernel}
\end{equation}
From this kernel, an incremental update for the $m$th particle of
$\mathbf{Z}$ at the $t$th step is computed by
\begin{equation}
\begin{aligned} & \phi_{t}^{\mathbf{Z}}\left(\mathbf{Z}_{t}^{\left(m\right)},\mathbf{\Theta}_{t}^{\left(m\right)}\right)\\
 & =\frac{1}{M}\sum_{r=1}^{M}\Bigg[k\left(\left(\mathbf{Z}_{t}^{\left(r\right)},\mathbf{\Theta}_{t}^{\left(r\right)}\right),\left(\mathbf{Z}_{t}^{\left(m\right)},\mathbf{\Theta}_{t}^{\left(m\right)}\right)\right)\\
 & \quad\cdot\nabla_{\mathbf{Z}_{t}^{\left(r\right)}}\log p\left(\mathbf{Z}_{t}^{\left(r\right)},\mathbf{\Theta}_{t}^{\left(r\right)}\mid\mathcal{D},\mathbf{P}\right)\\
 & \quad+\nabla_{\mathbf{Z}_{t}^{\left(r\right)}}k\left(\left(\mathbf{Z}_{t}^{\left(r\right)},\mathbf{\Theta}_{t}^{\left(r\right)}\right),\left(\mathbf{Z}_{t}^{\left(m\right)},\mathbf{\Theta}_{t}^{\left(m\right)}\right)\right)\Bigg].
\end{aligned}
\label{eq:update_z}
\end{equation}
A similar update for $\mathbf{\Theta}$ is proposed by replacing $\nabla_{\mathbf{Z}_{t}^{\left(r\right)}}$
by $\nabla_{\mathbf{\Theta}_{t}^{\left(r\right)}}$ as
\begin{equation}
\begin{aligned}\begin{aligned} & \phi_{t}^{\mathbf{\Theta}}\left(\mathbf{Z}_{t}^{\left(m\right)},\mathbf{\Theta}_{t}^{\left(m\right)}\right)\\
 & =\frac{1}{M}\sum_{r=1}^{M}\Bigg[k\left(\left(\mathbf{Z}_{t}^{\left(r\right)},\mathbf{\Theta}_{t}^{\left(r\right)}\right),\left(\mathbf{Z}_{t}^{\left(m\right)},\mathbf{\Theta}_{t}^{\left(m\right)}\right)\right)\\
 & \quad\cdot\nabla_{\mathbf{\Theta}_{t}^{\left(r\right)}}\log p\left(\mathbf{Z}_{t}^{\left(r\right)},\mathbf{\Theta}_{t}^{\left(r\right)}\mid\mathcal{D},\mathbf{P}\right)\\
 & \quad+\nabla_{\mathbf{\Theta}_{t}^{\left(r\right)}}k\left(\left(\mathbf{Z}_{t}^{\left(r\right)},\mathbf{\Theta}_{t}^{\left(r\right)}\right),\left(\mathbf{Z}_{t}^{\left(m\right)},\mathbf{\Theta}_{t}^{\left(m\right)}\right)\right)\Bigg].
\end{aligned}
\\
\\
\\
\end{aligned}
\label{eq:update_theta}
\end{equation}
The SVGD algorithm for inferring $p\left(\mathbf{G},\mathbf{\Theta}\mid\mathcal{D},\mathbf{P}\right)$
is represented in Algorithm~\ref{alg:SVGD}.

\section{Experiments\label{sec:Experiments}}

\begin{figure*}[t]
\begin{centering}
\includegraphics[width=1\textwidth]{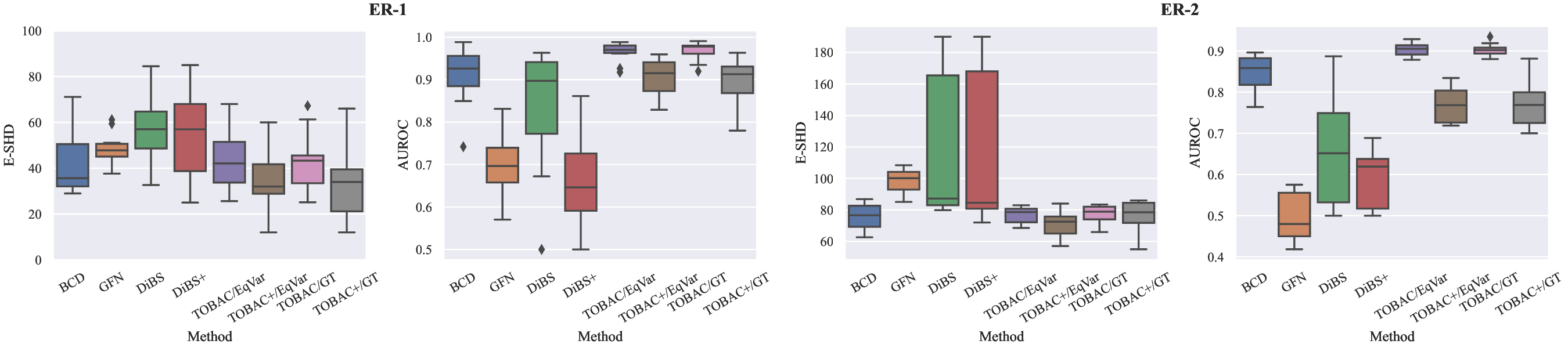}
\par\end{centering}
\caption{\label{fig:res_lin}Performance on synthetic data generated from linear
Gaussian models with $d=20$ variables, $n=100$ observations, and
$1000$ sampling steps. Lower E-SHD and higher AUROC are preferable.
Our TOBAC models with the orderings from EqVar~\cite{Chen_etal_19Causal}
and the ground-truth orderings are compared with BCD~Nets~(denoted
as BCD)~\cite{Cundy_etal_21BCD}, DAG-GFlowNet~(denoted as GFN)~\cite{Deleu_etal_22Bayesian},
and DiBS~\cite{Lorch_etal_21DiBS}. The DiBS+ and TOBAC+ models are
the results with weighted particle mixture. All the methods except
for DiBS are designed to ensure acyclicity, so the cyclicity score
is not necessary. The different designs of BCD~Nets and DAG-GFlowNet
make the comparison by the negative log likelihood evaluation with
the joint posterior distribution implausible. Our TOBAC models accomplish
the lowest E-SHD scores and the almost highest AUROC scores in both
ER-1 (sparse graph) and ER-2 (denser graph) settings.}
\end{figure*}

\begin{figure*}[t]
\centering{}\includegraphics[width=1\textwidth]{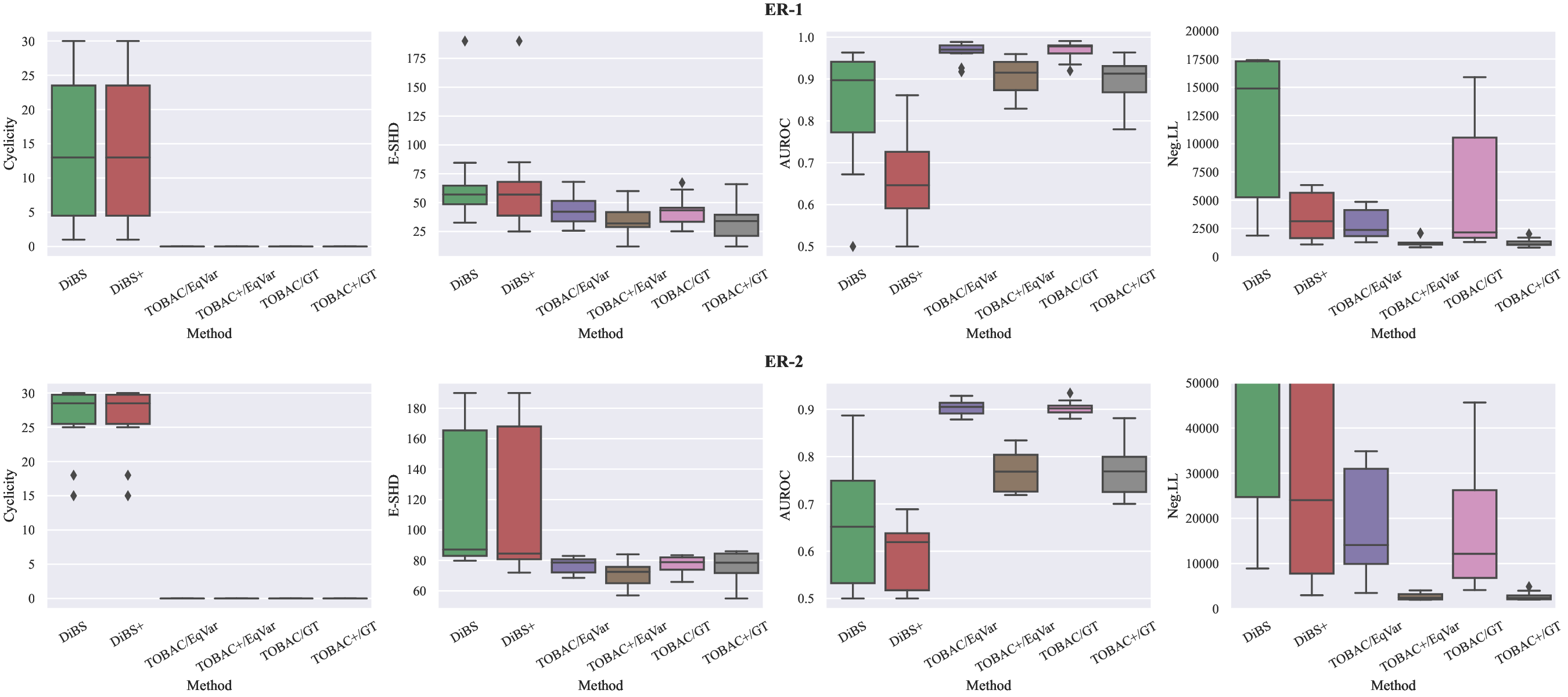}\caption{\label{fig:res_non}Performance on synthetic data generated from nonlinear
Gaussian models with $d=20$ variables, $n=100$ observations, and
$1000$ sampling steps. Lower Cyclicity, E-SHD, and Neg.LL, and higher
AUROC are preferable. Our TOBAC models with the orderings from EqVar~\cite{Chen_etal_19Causal}
and the ground-truth orderings are compared with DiBS~\cite{Lorch_etal_21DiBS}.
The DiBS+ and TOBAC+ models are the results with weighted particle
mixture. Note that BCD Nets~\cite{Cundy_etal_21BCD} and DAG-GFlowNet~\cite{Deleu_etal_22Bayesian}
are only designed for linear Gaussian models, so these methods are
not included in this experiment. In comparison with DiBS, our TOBAC
models can guarantee the acyclicty of the learned graphs. Our models
also perform better in all evaluation metrics and show more stable
results with fewer variances, especially in denser ER-2 graphs.}
\end{figure*}

\begin{figure*}[t]
\begin{centering}
\includegraphics[width=1\textwidth]{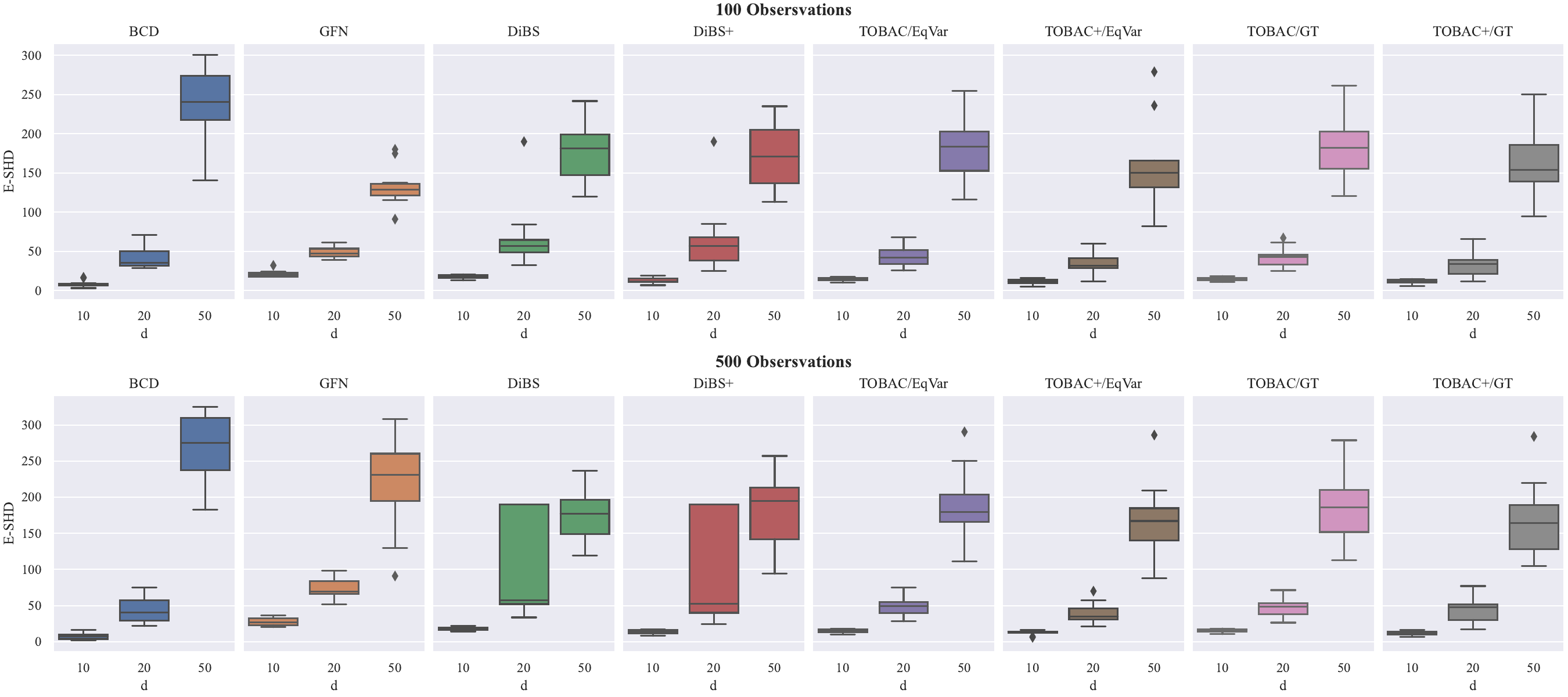}
\par\end{centering}
\caption{\label{fig:abl_n_d}Expected structural Hamming distance on synthetic
data with different sample sizes and dimensionalities. Lower E-SHD
is preferable. Our TOBAC models with the orderings from EqVar~\cite{Chen_etal_19Causal}
and the ground-truth orderings are compared with BCD~Nets~(denoted
as BCD)~\cite{Cundy_etal_21BCD}, DAG-GFlowNet~(denoted as GFN)~\cite{Deleu_etal_22Bayesian},
and DiBS~\cite{Lorch_etal_21DiBS}. The DiBS+ and TOBAC+ models are
the results with weighted particle mixture. Our models can learn more
accurate graphs with more stable results when the data's dimension
increases. Most approaches are not affected by the number of observations,
except for DAG-GFlowNet~(GFN)~\cite{Deleu_etal_22Bayesian}, which
cannot learn as effectively due to the incapability to handle higher
peakness of the posterior distribution.}
\end{figure*}

\begin{figure*}[t]
\begin{centering}
\includegraphics[width=1\textwidth]{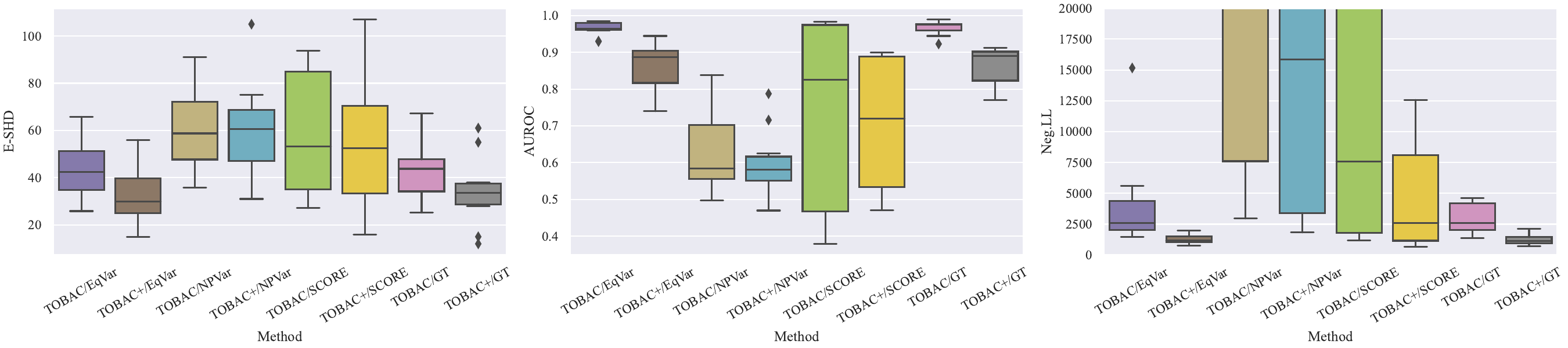}
\par\end{centering}
\caption{\label{fig:abl_ord}Performance with topological orderings from different
methods in the linear setting with $d=20$ variables, $n=100$ observations,
and $1000$ sampling steps. Lower E-SHD and Neg.LL, and higher AUROC
are preferable. The TOBAC models with the orderings from EqVar~\cite{Chen_etal_19Causal},
NPVar~\cite{Gao_etal_20Polynomial}, and SCORE~\cite{Rolland_etal_22Score}
are compared with the models with ground-truth orderings. The TOBAC+
models are the results with weighted particle mixture. The experiments
are configured with equal variance, so our models with orderings from
EqVar~\cite{Chen_etal_19Causal} can accomplish similar results to
the ones with the ground-truth orderings. The assumption of nonlinearity
in the setting of SCORE makes the models with its orderings less stable.}
\end{figure*}

\subsection{Experimental Settings\label{subsec:ExperimentalSettings}}

We compare our TOBAC approach with related Bayesian score-based methods
including BCD~Nets~\cite{Cundy_etal_21BCD}, DAG-GFlowNet~(GFN)~\cite{Deleu_etal_22Bayesian},
and DiBS~\cite{Lorch_etal_21DiBS} on synthetic and the flow cytometry
data~\cite{Sachs_etal_05Causal}. Beside DiBS, which can learn the
joint distribution $p\left(\mathbf{G},\mathbf{\Theta}\mid\mathcal{D}\right)$
and infer nonlinear Gaussian networks, BCD~Nets and DAG-GFlowNet
are designed to work with linear Gaussian models. BCD~Nets learns
the parameters using the weighted adjacency matrix and DAG-GFlowNet
uses the BGe score~\cite{Geiger_Heckerman_94Learning}. 

Regarding the selection of topological ordering for conditioning,
we choose the topological ordering from EqVar~\cite{Chen_etal_19Causal}
and the ground-truth~(GT) ordering as the condition for our approach.
We also analyze the effect of the ordering on the performance by replacing
the ordering with the one from NPVar~\cite{Gao_etal_20Polynomial}
and SCORE~\cite{Rolland_etal_22Score}. The DiBS+ and TOBAC+ denotations
in our experiments are the results with weighted particle mixture
in~\cite{Lorch_etal_21DiBS} that use $p\left(\mathbf{G},\mathbf{\Theta},\mathcal{D}\right)$
as the weight for each particle. This weight is employed as the unnormalized
probability for each inferred particle when computing the expectation
of the evaluation metrics.

\paragraph{Synthetic data \& graph prior}

We generate the data using the Erd\H{o}s\textendash R\'{e}nyi~(ER)
structure~\cite{Erdos_Renyi_60Evolution} with the degree of 1 and
2. In all settings in Section~\ref{subsec:ResultsSynthetic}, each
inference is performed on $n=100$ observations. For the ablation
study in Section~\ref{subsec:AblationStudy}, synthetic data with
$d\in\left\{ 10,20,50\right\} $ variables and $n\in\left\{ 100,500\right\} $
observations is utilized for the analysis of the effect of dimensionality
and sample size on the performance. For the graph prior, BCD~Nets
uses the Horseshoe prior for their strictly lower triangular matrix
$\mathbf{L}$, and GFN uses the prior from~\cite{Eggeling_etal_19Structure}.
DiBS and our approach use the prior of the Erd\H{o}s\textendash R\'{e}nyi
graphs, which is in the form of $p\left(\mathbf{G}\right)\propto q^{\left\Vert \mathbf{G}\right\Vert _{1}}\left(1-q\right)^{\binom{d}{2}-\left\Vert \mathbf{G}\right\Vert _{1}}$
where $q$ is the probability for an independent edge being added
to the DAG. 

\paragraph{Evaluation metrics}

Following the evaluation metrics used by previous work~\cite{Deleu_etal_22Bayesian,Cundy_etal_21BCD,Lorch_etal_21DiBS},
we evaluate the performance using the \textit{expected structural
Hamming distance} (E-SHD) and \textit{area under the receiver operating
characteristic curve} (AUROC). We follow \cite{Lorch_etal_21DiBS}
where the E-SHD score is the expectation of the \textit{structural
Hamming distance} (SHD) between each $\mathbf{G}$ and the ground-truth
$\mathbf{G}^{*}$ over the posterior distribution $p\left(\mathbf{G}\mid\mathcal{D}\right)$
to compute the expected number of edges that has been incorrectly
predicted. The formulation of this evaluation score is
\begin{align}
\text{E-SHD}\left(p,\mathbf{G}^{*}\right) & :=\sum_{\mathbf{G}}p\left(\mathbf{G}\mid\mathcal{D}\right)\mathrm{SHD}\left(\mathbf{G},\mathbf{G}^{*}\right).\label{eq:eshd}
\end{align}
The AUROC score is computed for each edge probability $p\left(G_{i,j}=1\mid\mathcal{D}\right)$
in comparison with the corresponding ground-truth edge in $\mathbf{G}^{*}$~\cite{Husmeier_03Sensitivity}.
In addition, we also evaluate the nonlinear Gaussian Bayesian networks
from DiBS and ours by the \textit{cyclicity} score and the average
\textit{negative log likelihood}~\cite{Lorch_etal_21DiBS}. The cyclicity
score is proposed by~\cite{Yu_etal_19DAG} and is used in DiBS as
the constraint for acyclicity. The score measuring the cyclicity or
non-DAG-ness of a graph $\mathbf{G}$ is defined as
\begin{equation}
h\left(\mathbf{G}\right):=\mathrm{tr}\left[\left(\mathbf{I}+\frac{1}{d}\mathbf{G}\right)^{d}\right]-d,\label{eq:cyclicity}
\end{equation}
where $h\left(\mathbf{G}\right)=0$ if and only if $\mathbf{G}$ has
no cycle and the higher its value, the more cyclic G becomes. In the
average negative log likelihood evaluation, a test dataset $\mathcal{D}^{test}$
containing 100 held-out observations are also generated to compute
the score as follows
\begin{equation}
\begin{aligned} & \textrm{Neg.LL}\left(p,\mathcal{D}^{test}\right)\\
 & :=-\sum_{\mathbf{G},\mathbf{\Theta}}p\left(\mathbf{G},\mathbf{\Theta}\mid\mathcal{D}\right)\log p\left(\mathcal{D}^{test}\mid\mathbf{G},\mathbf{\Theta}\right).
\end{aligned}
\label{eq:negll}
\end{equation}
This score is designed to evaluate the model's ability to predict
future observations. Note that BCD~Nets formulates the parameters
with the weighted adjacency matrix, and DAG-GFlowNet only estimates
the marginal posterior distribution $p\left(\mathbf{G}\mid\mathcal{D}\right)$
and uses BGe score~\cite{Geiger_Heckerman_94Learning} for the parameters.
Hence, we cannot compute the joint posterior distribution $p\left(\mathbf{G},\mathbf{\Theta}\mid\mathcal{D}\right)$
for the negative log likelihood evaluation. The reported results in
the following sections are obtained from ten randomly generated datasets
for each method and configuration.

\subsection{Performance on Synthetic Data\label{subsec:ResultsSynthetic}}

\paragraph{Linear Gaussian models}

Figure~\ref{fig:res_lin} illustrates the performance of the methods
with linear Gaussian models. We find that our TOBAC and TOBAC+ outperform
other approaches. TOBAC models accomplish the lowest E-SHD scores
and the almost highest AUROC scores in both ER-1 (sparse graph) and
ER-2 (denser graph) settings. In comparison with DiBS, the results
demonstrate that by introducing the prior knowledge, which is the
EqVar and ground-truth orderings, to the inference process can increase
the performance significantly. As the graph becomes denser, as in
the ER-2 settings, DiBS models have high variances in the results,
whereas our models are as stable as other approaches. Considering
DAG-GFlowNet, which uses GFlowNets~\cite{Bengio_etal_22GFlowNet}
to infer the posterior distribution, our models achieve lower E-SHD
score and substantially higher AUROC score in the denser graph setting.

\paragraph{Nonlinear Gaussian models}

The performance results of nonlinear Gaussian models are depicted
in Figure~\ref{fig:res_non}. The cyclicity scores clearly show that
the acyclicity constraint of DiBS is not as effective compared to
our approach. In addition to the certainty of acyclicity, from all
the E-SHD, AUROC, and negative log likelihood scores, we can see that
our approaches can infer better graphs than DiBS. This improvement
in performance emphasizes the benefit of the  orderings while guaranteeing
the acyclicity at every number of sampling steps. With the  orderings
given, optimizing the graph parameters become significantly easier.
As a result, the log likelihood of the observational data is improved.
Furthermore, the plots in the ER-2 indicate that TOBAC is more stable
than DiBS when inferring denser graphs. 

\subsection{Performance on Real Data\label{subsec:ResultsSachs}}

\begin{table}[t]
\caption{\label{tab:res_sachs}Performance on the flow cytometry dataset~\cite{Sachs_etal_05Causal}.
\\Reported results are the mean$\pm$std of the metrics. \\The best
results are presented in \textbf{bold} style. \\Our methods achieve
the best E-SHD score and the second highest AUROC score in comparison
with related approaches.}

\begin{centering}
\begin{tabular}{lcc}
\toprule 
 & E-SHD$\downarrow$ & AUROC$\uparrow$\tabularnewline
\midrule
BCD~\cite{Cundy_etal_21BCD} & $17.1\pm0.13$ & $0.534\pm0.0640$\tabularnewline
GFN~\cite{Deleu_etal_22Bayesian} & $21.2\pm1.50$ & $0.496\pm0.0609$\tabularnewline
DiBS~\cite{Lorch_etal_21DiBS} & $15.9\pm0.41$ & $\mathbf{0.625\pm0.0367}$\tabularnewline
DiBS+~\cite{Lorch_etal_21DiBS} & $15.4\pm1.35$ & $0.553\pm0.0494$\tabularnewline
TOBAC/EqVar & $15.6\pm0.42$ & $0.599\pm0.0395$\tabularnewline
TOBAC+/EqVar & \textbf{$\mathbf{14.8\pm1.13}$} & $0.560\pm0.0309$\tabularnewline
\bottomrule
\end{tabular}
\par\end{centering}
\vspace{-3em}
\end{table}

The flow cytometry dataset includes $n=853$ observational continuous
data points and $d=11$ phosphoproteins. The  graphs are inferred
with linear Gaussian models and Erd\H{o}s\textendash R\'{e}nyi graph
priors at $1000$ sampling steps. As we can observe in Table~\ref{tab:res_sachs},
our TOBAC and TOBAC+ approaches with the orderings from EqVar has
achieved the lowest E-SHD values at $15.6\pm0.42$ and $14.8\pm1.13$
respectively. Additionally, the AUROC score of TOBAC is the second
highest one at $0.599\pm0.0395$, which is lower than DiBS's at $0.625\pm0.0367$.
Moreover, TOBAC+'s AUROC score is slightly higher than DiBS+'s score
at $0.560\pm0.0309$ compared to $0.553\pm0.0494$.

\subsection{Ablation Study\label{subsec:AblationStudy}}

\paragraph{Sample sizes \& dimensionalities}

We study the effect of different sample sizes and data dimensionalities
on TOBAC and related approaches. As we can perceive in Figure~\ref{fig:abl_n_d},
the number of observations does not have much effect on the performance
in most cases. In the case of GFN, as also described by Deleu et al.~\cite{Deleu_etal_22Bayesian},
when the number of observations increases, this model can not perform
as well as with a smaller sample size due to the higher peakness of
the posterior distribution. These results show that Bayesian approaches
can sufficiently handle the uncertainty caused by a smaller amount
of data. In contrast, as the number of dimensions in the data rises,
the effect on the inference performance diverges among the approaches.
Graphs with 50 nodes inferred by TOBAC are more accurate in comparison
with other approaches. 

\paragraph{Topological orderings from other approaches}

We summarize the results of the graphs inferred by TOBAC with the
ground-truth ordering and the orderings learned by several approaches
consisting of EqVar~\cite{Chen_etal_19Causal}, NPVar~\cite{Gao_etal_20Polynomial},
and SCORE \cite{Rolland_etal_22Score} in Figure~\ref{fig:abl_ord}.
In this setting, the variances of the variables are equal, which are
matched to EqVar settings. As a consequence, the results suggest that
TOBAC with EqVar can have the performance scores that are close to
TOBAC with the ground-truth ordering on synthetic data. Although being
the state-of-the-art approach in learning the ordering, the results
when employing SCORE are more unstable in comparison with other simpler
ones. This unstable may be due to the nonlinear assumption of the
structural equation model in SCORE, which does not fit well with this
linear Gaussian setting.

\paragraph{Uniform \& weighted particle mixture}

From the observed results, the weighted particle mixture can reduce
the E-SHD and Neg.LL scores. However, it will also cause a reduction
in the AUROC scores. The lower performance of the uniform mixture
can be due to the crude approximation of the particles' posterior
probability mass function, which is replaced by a more meaningful
approach in the weighted mixture~\cite{Lorch_etal_21DiBS}.

\section{Conclusion\label{sec:Conclusion}}

In this work, we have presented TOBAC\textemdash an approach to strictly
constraining the acyclicity of the inferred graphs and integrating
the knowledge from the topological orderings into the inference process.
Our work uses a continuous representation of the canonical adjacency
matrix and approximate the posterior using particle variational inference.
Our proposed framework makes the inference process less complicated
and enhances the accuracy of inferred graphs and parameters. Accordingly,
our work can outperform most related Bayesian score-based approaches
on both synthetic and real observational data. In future work, we
will explore methods to learn the posterior distribution of the topological
orderings with observational data, which will allow our approach to
infer more diverse posterior distributions while still ensuring the
acyclicity of generated graphs.

\bibliographystyle{IEEEtranS}
\bibliography{IEEEabrv,icdm2023_tobac}

\appendices{}

\section{Proof of Equation~\eqref{eq:expectation}}

First, we will recall the generative model in Equation~(\ref{eq:gen_factorize}),
which states that 
\[
\begin{aligned} & p\left(\mathbf{Z},\mathbf{S},\mathbf{G},\mathbf{\Theta},\mathcal{D}\mid\mathbf{P}\right)=\\
 & p\left(\mathbf{Z}\right)p\left(\mathbf{S}\mid\mathbf{Z}\right)p\left(\mathbf{G}\mid\mathbf{S},\mathbf{P}\right)p\left(\mathbf{\Theta}\mid\mathbf{G}\right)p\left(\mathcal{D}\mid\mathbf{G},\mathbf{\Theta}\right)
\end{aligned}
\]
In addition, from Bayes' theorem, we have 
\[
p\left(\mathbf{Z}\right)=\frac{p\left(\mathcal{D}\right)p\left(\mathbf{Z},\Theta\mid\mathcal{D},\mathbf{P}\right)}{p\left(\Theta,\mathcal{D}\mid\mathbf{Z},\mathbf{P}\right)}\Rightarrow\frac{p\left(\mathbf{Z}\right)}{p\left(\mathcal{D}\right)}=\frac{p\left(\mathbf{Z},\Theta\mid\mathcal{D},\mathbf{P}\right)}{p\left(\Theta,\mathcal{D}\mid\mathbf{Z},\mathbf{P}\right)}.
\]

\begin{IEEEproof}
The proof of the computation for expectation of $f\left(\mathbf{G},\mathbf{\Theta}\right)$
with $\left(\mathbf{G},\mathbf{\Theta}\right)\sim p\left(\mathbf{G},\mathbf{\Theta}\mid\mathcal{D},\mathbf{P}\right)$
in Equation~(\ref{eq:expectation}) is as follows
\begin{align*}
 & \mathbb{E}_{p\left(\mathbf{G},\mathbf{\Theta}\mid\mathcal{D},\mathbf{P}\right)}\left[f\left(\mathbf{G},\mathbf{\Theta}\right)\right]\\
 & =\sum_{\mathbf{G}}\int_{\mathbf{\Theta}}p\left(\mathbf{G},\mathbf{\Theta}\mid\mathcal{D},\mathbf{P}\right)f\left(\mathbf{G},\mathbf{\Theta}\right)d\mathbf{\Theta}\\
 & =\sum_{\mathbf{G}}\int_{\mathbf{\Theta}}\frac{p\left(\mathbf{G},\mathbf{\Theta},\mathcal{D}\mid\mathbf{P}\right)f\left(\mathbf{G},\mathbf{\Theta}\right)}{p\left(\mathcal{D}\right)}d\mathbf{\Theta}\\
 & =\sum_{\mathbf{G}}\int_{\mathbf{\Theta}}\sum_{\mathbf{S}}\int_{\mathbf{Z}}\frac{p\left(\mathbf{Z},\mathbf{S},\mathbf{G},\mathbf{\Theta},\mathcal{D}\mid\mathbf{P}\right)f\left(\mathbf{G},\mathbf{\Theta}\right)}{p\left(\mathcal{D}\right)}d\mathbf{Z}d\mathbf{\Theta}\\
 & =\sum_{\mathbf{G}}\int_{\mathbf{\Theta}}\sum_{\mathbf{S}}\int_{\mathbf{Z}}\frac{1}{p\left(\mathcal{D}\right)}p\left(\mathbf{Z}\right)p\left(\mathbf{S}\mid\mathbf{Z}\right)p\left(\mathbf{G}\mid\mathbf{S},\mathbf{P}\right)\cdot\\
 & \qquad p\left(\mathbf{\Theta}\mid\mathbf{G}\right)p\left(\mathcal{D}\mid\mathbf{G},\mathbf{\Theta}\right)f\left(\mathbf{G},\mathbf{\Theta}\right)d\mathbf{Z}d\mathbf{\Theta}\\
 & =\int_{\mathbf{\Theta}}\int_{\mathbf{Z}}\frac{p\left(\mathbf{Z}\right)}{p\left(\mathcal{D}\right)}\sum_{\mathbf{G}}\sum_{\mathbf{S}}p\left(\mathbf{S}\mid\mathbf{Z}\right)p\left(\mathbf{G}\mid\mathbf{S},\mathbf{P}\right)\cdot\\
 & \qquad p\left(\mathbf{\Theta}\mid\mathbf{G}\right)p\left(\mathcal{D}\mid\mathbf{G},\mathbf{\Theta}\right)f\left(\mathbf{G},\mathbf{\Theta}\right)d\mathbf{Z}d\mathbf{\Theta}\\
 & =\int_{\mathbf{\Theta}}\int_{\mathbf{Z}}\frac{p\left(\mathbf{Z}\right)}{p\left(\mathcal{D}\right)}\sum_{\mathbf{G}}\sum_{\mathbf{S}}p\left(\mathbf{G},\mathbf{S}\mid\mathbf{Z},\mathbf{P}\right)\cdot\\
 & \qquad p\left(\mathbf{\Theta}\mid\mathbf{G}\right)p\left(\mathcal{D}\mid\mathbf{G},\mathbf{\Theta}\right)f\left(\mathbf{G},\mathbf{\Theta}\right)d\mathbf{Z}d\mathbf{\Theta}\\
 & =\int_{\mathbf{\Theta}}\int_{\mathbf{Z}}\frac{p\left(\mathbf{Z}\right)}{p\left(\mathcal{D}\right)}\sum_{\mathbf{G}}p\left(\mathbf{G}\mid\mathbf{Z},\mathbf{P}\right)\cdot\\
 & \qquad p\left(\mathbf{\Theta}\mid\mathbf{G}\right)p\left(\mathcal{D}\mid\mathbf{G},\mathbf{\Theta}\right)f\left(\mathbf{G},\mathbf{\Theta}\right)d\mathbf{Z}d\mathbf{\Theta}\\
 & =\int_{\mathbf{\Theta}}\int_{\mathbf{Z}}\frac{p\left(\mathbf{Z},\mathbf{\Theta}\mid\mathcal{D},\mathbf{P}\right)}{p\left(\mathcal{D},\mathbf{\Theta}\mid\mathbf{Z},\mathbf{P}\right)}\sum_{\mathbf{G}}p\left(\mathbf{G}\mid\mathbf{Z},\mathbf{P}\right)\cdot\\
 & \qquad p\left(\mathbf{\Theta}\mid\mathbf{G}\right)p\left(\mathcal{D}\mid\mathbf{G},\mathbf{\Theta}\right)f\left(\mathbf{G},\mathbf{\Theta}\right)d\mathbf{Z}d\mathbf{\Theta}\\
 & =\int_{\mathbf{\Theta}}\int_{\mathbf{Z}}p\left(\mathbf{Z},\mathbf{\Theta}\mid\mathcal{D},\mathbf{P}\right)\frac{1}{p\left(\mathcal{D},\mathbf{\Theta}\mid\mathbf{Z},\mathbf{P}\right)}\sum_{\mathbf{G}}p\left(\mathbf{G}\mid\mathbf{Z},\mathbf{P}\right)\cdot\\
 & \qquad p\left(\mathbf{\Theta}\mid\mathbf{G}\right)p\left(\mathcal{D}\mid\mathbf{G},\mathbf{\Theta}\right)f\left(\mathbf{G},\mathbf{\Theta}\right)d\mathbf{Z}d\mathbf{\Theta}\\
 & =\int_{\mathbf{\Theta}}\int_{\mathbf{Z}}p\left(\mathbf{Z},\mathbf{\Theta}\mid\mathcal{D},\mathbf{P}\right)\frac{1}{\sum_{\mathbf{G}}p\left(\mathcal{D},\mathbf{G},\mathbf{\Theta}\mid\mathbf{Z},\mathbf{P}\right)}\cdot\\
 & \qquad\sum_{\mathbf{G}}p\left(\mathbf{G}\mid\mathbf{Z},\mathbf{P}\right)p\left(\mathbf{\Theta}\mid\mathbf{G}\right)\cdot\\
 & \qquad\qquad p\left(\mathcal{D}\mid\mathbf{G},\mathbf{\Theta}\right)f\left(\mathbf{G},\mathbf{\Theta}\right)d\mathbf{Z}d\mathbf{\Theta}\\
 & =\int_{\mathbf{\Theta}}\int_{\mathbf{Z}}p\left(\mathbf{Z},\mathbf{\Theta}\mid\mathcal{D},\mathbf{P}\right)\cdot\\
 & \qquad\frac{\sum_{\mathbf{G}}p\left(\mathbf{G}\mid\mathbf{Z},\mathbf{P}\right)p\left(\mathbf{\Theta}\mid\mathbf{G}\right)p\left(\mathcal{D}\mid\mathbf{G},\mathbf{\Theta}\right)f\left(\mathbf{G},\mathbf{\Theta}\right)}{\sum_{\mathbf{G}}p\left(\mathbf{G}\mid\mathbf{Z},\mathbf{P}\right)p\left(\mathbf{\Theta}\mid\mathbf{G}\right)p\left(\mathcal{D}\mid\mathbf{G},\mathbf{\Theta}\right)}\cdot\\
 & \qquad\qquad d\mathbf{Z}d\mathbf{\Theta}\\
 & =\int_{\mathbf{\Theta}}\int_{\mathbf{Z}}p\left(\mathbf{Z},\mathbf{\Theta}\mid\mathcal{D},\mathbf{P}\right)\cdot\\
 & \qquad\frac{\sum_{\mathbf{G}}p\left(\mathbf{G}\mid\mathbf{Z},\mathbf{P}\right)p\left(\mathbf{\Theta},\mathcal{D}\mid\mathbf{G}\right)f\left(\mathbf{G},\mathbf{\Theta}\right)}{\sum_{\mathbf{G}}p\left(\mathbf{G}\mid\mathbf{Z},\mathbf{P}\right)p\left(\mathbf{\Theta},\mathcal{D}\mid\mathbf{G}\right)}d\mathbf{Z}d\mathbf{\Theta}\\
 & =\mathbb{E}_{p\left(\mathbf{Z},\mathbf{\Theta}\mid\mathcal{D},\mathbf{P}\right)}\left[\frac{\mathbb{E}_{p\left(\mathbf{G}\mid\mathbf{Z},\mathbf{P}\right)}\left[f\left(\mathbf{G},\mathbf{\Theta}\right)p\left(\mathbf{\Theta},\mathcal{D}\mid\mathbf{G}\right)\right]}{\mathbb{E}_{p\left(\mathbf{G}\mid\mathbf{Z},\mathbf{P}\right)}\left[p\left(\mathbf{\Theta},\mathcal{D}\mid\mathbf{G}\right)\right]}\right].
\end{align*}
\end{IEEEproof}

\section{Proof of Equations~\eqref{eq:grad_z} \&~\eqref{eq:grad_theta}}

\begin{IEEEproof}
The derivative of the posterior log likelihood $\log p\left(\mathbf{Z},\mathbf{\Theta}\mid\mathcal{D},\mathbf{P}\right)$
with respect to the latent variable $\mathbf{Z}$ in Equation~(\ref{eq:grad_z})
is computed as
\begin{align*}
 & \nabla_{\mathbf{Z}}\log p\left(\mathbf{Z},\mathbf{\Theta}\mid\mathcal{D},\mathbf{P}\right)\\
 & =\nabla_{\mathbf{Z}}\log p\left(\mathbf{Z},\mathbf{\Theta},\mathcal{D}\mid\mathbf{P}\right)-\nabla_{\mathbf{Z}}\log p\left(\mathcal{D}\mid\mathbf{P}\right)\\
 & =\nabla_{\mathbf{Z}}\log p\left(\mathbf{Z},\mathbf{\Theta},\mathcal{D}\mid\mathbf{P}\right)\quad\textrm{(as }\nabla_{\mathbf{Z}}\log p\left(\mathcal{D}\mid\mathbf{P}\right)=0\textrm{)}\\
 & =\nabla_{\mathbf{Z}}\log p\left(\mathbf{Z}\right)+\nabla_{\mathbf{Z}}\log p\left(\mathbf{\Theta},\mathcal{D}\mid\mathbf{Z},\mathbf{P}\right)\\
 & =\nabla_{\mathbf{Z}}\log p\left(\mathbf{Z}\right)+\frac{\nabla_{\mathbf{Z}}p\left(\mathbf{\Theta},\mathcal{D}\mid\mathbf{Z},\mathbf{P}\right)}{p\left(\mathbf{\Theta},\mathcal{D}\mid\mathbf{Z},\mathbf{P}\right)}\\
 & \qquad\qquad\textrm{(as }\nabla_{\mathbf{x}}\log g\left(\mathbf{x}\right)=\nabla_{\mathbf{x}}g\left(\mathbf{x}\right)/g\left(\mathbf{x}\right)\textrm{)}\\
 & =\nabla_{\mathbf{Z}}\log p\left(\mathbf{Z}\right)+\frac{\nabla_{\mathbf{Z}}\left[\sum_{G}p\left(\mathbf{G},\mathbf{\Theta},\mathcal{D}\mid\mathbf{Z},\mathbf{P}\right)\right]}{\sum_{G}p\left(\mathbf{G},\mathbf{\Theta},\mathcal{D}\mid\mathbf{Z},\mathbf{P}\right)}\\
 & =\nabla_{\mathbf{Z}}\log p\left(\mathbf{Z}\right)+\frac{\nabla_{\mathbf{Z}}\left[\sum_{G}p\left(\mathbf{G}\mid\mathbf{Z},\mathbf{P}\right)p\left(\mathbf{\Theta},\mathcal{D}\mid\mathbf{G}\right)\right]}{\sum_{G}p\left(\mathbf{G}\mid\mathbf{Z},\mathbf{P}\right)p\left(\mathbf{\Theta},\mathcal{D}\mid\mathbf{G}\right)}\\
 & =\nabla_{\mathbf{Z}}\log p\left(\mathbf{Z}\right)+\frac{\nabla_{\mathbf{Z}}\mathbb{E}_{p\left(\mathbf{G}\mid\mathbf{Z},\mathbf{P}\right)}\left[p\left(\mathbf{\Theta},\mathcal{D}\mid\mathbf{G}\right)\right]}{\mathbb{E}_{p\left(\mathbf{G}\mid\mathbf{Z},\mathbf{P}\right)}\left[p\left(\mathbf{\Theta},\mathcal{D}\mid\mathbf{G}\right)\right]}.
\end{align*}
The derivative of the posterior log likelihood $\log p\left(\mathbf{Z},\mathbf{\Theta}\mid\mathcal{D},\mathbf{P}\right)$
with respect to the latent variable $\mathbf{\Theta}$ in Equation~(\ref{eq:grad_theta})
can also be computed easily by following these steps.

\end{IEEEproof}

\section{Implementation \& Hyperparameters}

TOBAC\footnote{\url{https://github.com/quangdzuytran/TOBAC}} is implemented
using \texttt{Python} and the \texttt{JAX}~\cite{jax2018github}
library. Our implementation is based on the implementation of DiBS~\cite{Lorch_etal_21DiBS}\footnote{\url{https://github.com/larslorch/dibs}}.
For other approaches including BCD~Nets~\cite{Cundy_etal_21BCD}\footnote{\url{https://github.com/ermongroup/BCD-Nets}}
and DAG-GFlowNet~\cite{Deleu_etal_22Bayesian}\footnote{\url{https://github.com/tristandeleu/jax-dag-gflownet}},
we use their original implementation for the experiments. 

For synthesizing the data, we follow previous approaches~\cite{Cundy_etal_21BCD,Lorch_etal_21DiBS,Deleu_etal_22Bayesian}
for data configurations. In the linear Gaussian synthetic data, the
parameters $\theta_{i}$ of the edges are sampled from $\mathcal{N}\left(0,1\right)$
with the additive noises sampled from $\mathcal{N}\left(0,0.1\right)$.
As in BCD~Nets~\cite{Cundy_etal_21BCD} and DiBS~\cite{Lorch_etal_21DiBS},
a minimum value of $0.5$ is applied to the parameters by adding $\operatorname{sign}\left(\theta_{i}\right)\times0.5$
to the sampled parameters to clarify whether the parent nodes have
contributions in the variable. This value is also chosen as the threshold
for determining if an edge exists in BCD~Nets. In nonlinear Gaussian
data, the data is generated from neural networks with one hidden layer
comprising of 5 nodes and \texttt{ReLU} activation. The parameters
in the neural networks are also sampled from $\mathcal{N}\left(0,1\right)$
with observation noises sampled from $\mathcal{N}\left(0,0.1\right)$
as in the linear Gaussian setting. These configurations are also applied
to the corresponding inference models. In the experiments with the
flow cytometry dataset~\cite{Sachs_etal_05Causal}, the linear Gaussian
model is chosen for inferring the probability of the parameters given
the graph. 

The ground-truth topological orderings in the experiments are obtained
from the ground-truth DAGs using the topological sorting algorithm~\cite{Manber_89Introduction}
of the \texttt{NetworkX}~\cite{Hagberg_etal_08Exploring} library.
In order to use EqVar~\cite{Chen_etal_19Causal}\footnote{\url{https://github.com/WY-Chen/EqVarDAG}}
and NPVar~\cite{Gao_etal_20Polynomial}\footnote{\url{https://github.com/MingGao97/NPVAR}}
for finding the orderings, we re-implement these methods with \texttt{NumPy}~\cite{Harris_etal_20Array}
and \texttt{pyGAM}~\cite{Serven_Brummitt_18pyGAM} from their \texttt{R}
implementations. Regarding SCORE~\cite{Rolland_etal_22Score}\footnote{\url{https://github.com/paulrolland1307/SCORE}},
as the \texttt{PyTorch}~\cite{Paszke_etal_19PyTorch} version of
the code is available, we utilize their original implementation.

All of the baselines are also implemented with the \texttt{JAX} library.
Our experiments with TOBAC and DiBS are performed with the hyperparameters
in Table~\ref{tab:hyperparameters}, which are chosen in accordance
to a similar approach of DiBS~\cite{Lorch_etal_21DiBS}.

\begin{table}[H]
\begin{centering}
\caption{\label{tab:hyperparameters}Chosen hyperparameters for TOBAC in the
experiments}
\emph{}%
\begin{tabular}{>{\centering}p{0.2\columnwidth}>{\raggedright}p{0.55\columnwidth}>{\centering}p{0.1\columnwidth}}
\toprule 
Hyperparameter & Description & Value\tabularnewline
\midrule
\midrule 
$M$ & Number of particles in Stein variational gradient descent & $30$\tabularnewline
\midrule 
$\tilde{\alpha}$ & The linear rate of $\alpha_{t}:=\tilde{\alpha}t$ & $0.05$\tabularnewline
\midrule 
$\left(\gamma_{z},\gamma_{\theta}\right)$ & Bandwidths of the kernel & $\left(5,500\right)$\tabularnewline
\midrule 
$\sigma_{z}$ & Standard deviation of latent variable $\mathbf{Z}$ & $1/\sqrt{k}$\tabularnewline
\midrule 
$\sigma_{obs}$ & Standard deviation of observation noise & $0.1$\tabularnewline
\midrule 
$\eta_{t_{0}}$ & Initial learning rate of RMSProp & $0.005$\tabularnewline
\midrule 
$\left(\mu_{e},\sigma_{e}\right)$ & Mean and standard deviation of the parameters in linear Gaussian models & $\left(0,1\right)$\tabularnewline
\midrule 
$\sigma_{p}$ & Standard deviation of the parameters in nonlinear Gaussian models & $1$\tabularnewline
\midrule 
$L$ & Number of hidden layers in nonlinear Gaussian models & $5$\tabularnewline
\bottomrule
\end{tabular}
\par\end{centering}
\end{table}

\balance
\end{document}